\documentclass[sigconf, nonacm]{acmart}

\usepackage{booktabs}
\usepackage{amsmath}
\usepackage{tabularx}
\usepackage{array}
\usepackage{xspace}
\usepackage{enumitem}
\usepackage{etoolbox}
\usepackage{tikz}
\usepackage{dblfloatfix}
\usepackage{placeins}
\usepackage{caption}
\newcolumntype{C}[1]{>{\centering\arraybackslash}p{#1}}
\usetikzlibrary{arrows.meta,positioning,shapes.geometric,fit}

\newcommand{\qrs}{\textsc{QRS}\xspace}

\newcommand{\nisq}{\textsc{NISQ}\xspace}
\newcolumntype{Y}{>{\raggedright\arraybackslash}X}
\begin{document}

\title[IonSense-QKG]{IonSense-QKG: A Quantum-Readiness Metadata Framework for Lithium-Ion Battery Dataset Discovery}

\author{Sakthi Prabhu Gunasekar}
\orcid{0009-0006-0153-5674}
\affiliation{%
  \institution{Amrita School of Computing, Amrita Vishwa Vidyapeetham}
  \city{Coimbatore}
  \country{India}
}
\email{sakthiprabhu.g@gmail.com}

\author{Prasanna Kumar Rangarajan}
\orcid{0000-0001-6103-259X}
\affiliation{%
  \institution{Amrita School of Computing, Amrita Vishwa Vidyapeetham}
  \city{Coimbatore}
  \country{India}
}
\email{r\_prasannakumar@cb.amrita.edu}

\begin{abstract}
Public lithium-ion battery datasets support state-of-health estimation,
remaining-useful-life prediction, anomaly detection, electrochemical
characterisation, second-life analytics, and battery-safety research. However,
these datasets remain difficult to reuse for near-term hybrid
quantum--classical machine learning because they vary widely in chemistry,
modality, scale, label quality, access status, sequence structure, and
preprocessing complexity. This paper presents \textbf{IonSense-QKG}, a
quantum-readiness metadata framework for lithium-ion battery dataset
discovery. Starting from EV-Battery-IonSense, we enrich public battery dataset
metadata with quantum-relevant fields including task, modality, chemistry,
label type, sequence type, access status, preprocessing needs, estimated qubit
range, and candidate quantum encodings. We define a weighted
\textbf{Quantum Readiness Score} (\qrs) for ranking datasets by practical
\nisq-era workflow feasibility. \qrs is not a claim of quantum advantage and
does not imply that quantum models were trained on all datasets; it is a
transparent prioritisation heuristic for future hybrid quantum benchmarks. We
demonstrate SQL-style discovery queries and release the metadata, scoring
script, ranked table, robustness outputs, link-checking script, and query
workload as an open artifact.
\end{abstract}

\maketitle

\vspace{.3cm}
\noindent\textbf{Artifact Availability.}
The source code, metadata schema, enriched annotations, Quantum Readiness Score
computation script, ranked metadata table, robustness outputs, link-checking
script, and query workload are available at
\url{https://github.com/SakthiGs/EV-Battery-IonSense}.

\section{Introduction}
\label{sec:intro}

Battery analytics has become a central problem in electric mobility, renewable energy storage, battery second life, and safety-critical energy systems. Public datasets now cover state-of-health (SoH) estimation, state-of-charge (SoC) estimation, remaining useful life (RUL) prediction, thermal fault detection, impedance-based diagnostics, degradation mode analysis, field-usage modelling, and battery imaging. In parallel, hybrid quantum--classical machine learning is increasingly explored for scientific learning tasks where nonlinear feature maps, kernel-based similarity, and parameter-efficient models may be useful under constrained data regimes~\cite{biamonte2017qml,schuld2019feature,cerezo2021vqa,havlicek2019supervised}.

Despite these parallel developments, a practical barrier remains: \emph{not every battery dataset is suitable for near-term quantum machine learning}. Some public datasets are extremely large, high-dimensional, irregularly sampled, image-heavy, weakly labelled, or difficult to access. Others are compact, well-labelled, and naturally suited to low-dimensional feature maps, short time-series encodings, or quantum-kernel benchmarking. Electrochemical impedance spectroscopy (EIS), for example, can often be represented as compact frequency-domain signatures, while fleet-scale EV telemetry may contain millions or billions of rows that require windowing, aggregation, or representation learning before any quantum encoding is feasible.

Existing battery dataset repositories are valuable navigational indexes, but they rarely expose metadata needed by quantum computing researchers: estimated qubit requirements, sequence-window feasibility, candidate encoding strategies, circuit-depth implications, or \nisq suitability. As a result, researchers must manually inspect datasets, infer task compatibility, and estimate whether a dataset can support a feasible hybrid quantum workflow. This paper addresses that gap by proposing \textbf{IonSense-QKG}, a quantum-readiness metadata framework for lithium-ion battery dataset discovery. The framework builds on EV-Battery-IonSense~\cite{ionsense}, a curated open-source index of public battery datasets and papers, and enriches it with metadata designed for hybrid quantum--classical analytics.

The goal is not to claim quantum advantage, nor to introduce a new battery dataset. Instead, we ask a data-management question: \emph{Can a queryable metadata framework help researchers identify public lithium-ion battery datasets that are feasible, relevant, and benchmark-ready for \nisq-era hybrid quantum machine learning?}

\noindent
We make three contributions:
\begin{enumerate}[leftmargin=*]
  \item \textbf{A quantum-ready metadata schema for battery dataset discovery.} We extend a curated lithium-ion battery dataset index with knowledge-graph-compatible metadata fields covering tasks, modalities, labels, sequence types, access status, preprocessing requirements, and candidate quantum encodings.
  \item \textbf{A heuristic Quantum Readiness Score for candidate selection.} We define a transparent scoring rubric that estimates the practical feasibility of using a dataset in \nisq-era hybrid quantum--classical experiments. The score is intended for dataset prioritisation, not as evidence of quantum advantage.
  \item \textbf{A query workload for quantum battery dataset discovery.} We demonstrate SQL-style queries over enriched metadata to identify datasets suitable for \nisq-feasible hybrid QML, quantum time-series encoding, and limited-label anomaly/failure detection.
\end{enumerate}

By positioning dataset selection as a knowledge-management problem, IonSense-QKG supports reproducible and transparent quantum battery analytics while remaining distinct from dataset-descriptor work and model-benchmark papers.

\section{Background and Motivation}
\label{sec:background}

\subsection{Battery datasets are heterogeneous}

Open battery datasets vary substantially in experimental setting, scale,
chemistry, modality, and annotation quality. EV-Battery-IonSense organises
public resources into multiple battery-data categories: (i) field EV
telemetry and charging data; (ii) lab ageing and degradation datasets;
(iii) safety, failure, and thermal-runaway resources; (iv) SoH/SoC/RUL
state-estimation datasets; (v) electrochemical characterisation such as
EIS, OCV, and impedance; (vi) battery imaging and ultrasound resources;
(vii) second-life and grid-storage datasets; and (viii) aerospace or
specialised battery applications~\cite{ionsense}. These categories differ
strongly in scale, modality, label quality, sequence structure, and
preprocessing feasibility, motivating a quantum-readiness layer rather
than a flat dataset index.

Field telemetry datasets may contain voltage, current, temperature, power,
and charging behaviour at high temporal resolution. Lab ageing datasets may
provide reference performance tests, charge/discharge curves, capacity
trajectories, hybrid pulse power characterisation, or impedance
measurements. Safety datasets may include rare failure labels, abuse tests,
internal short-circuit experiments, or surface strain measurements. Imaging
datasets may include CT, X-ray, synchrotron, or ultrasonic measurements
across spatial scales.

This heterogeneity is valuable for battery research but challenging for
quantum workflow design. A dataset that is excellent for classical deep
learning may be poorly suited for direct quantum encoding, while a smaller
electrochemical dataset may be more realistic for \nisq benchmarking.

\subsection{Dataset discovery as quantum data management}
Dataset discovery is usually treated as a manual literature-review task. Researchers search papers, inspect repositories, read data descriptors, and decide whether a dataset is suitable. Hybrid quantum--classical battery analytics introduces additional constraints: How many features must be encoded into a quantum circuit? Is the signal tabular, time-series, spectral, image-based, or graph-structured? Can the sequence be windowed or compressed into a short input vector? Are labels available for SoH, SoC, RUL, failure, thermal runaway, or anomaly detection? Is the dataset publicly accessible with sufficient documentation? What encoding strategy is plausible: angle encoding, amplitude encoding, quantum kernel, quantum reservoir, or variational quantum head?

These questions are not purely battery-domain questions; they are metadata, indexing, retrieval, provenance, and knowledge-representation questions. This places the problem within the scope of quantum computing and data/knowledge management.

\subsection{Why quantum readiness matters}
Near-term quantum devices have limited qubit counts, noisy gates, finite shot budgets, and circuit-depth constraints. Even when using simulators, practical hybrid quantum ML workflows require feature selection, dimensionality reduction, and encoding design. Therefore, not all datasets should be treated equally. A billion-row field telemetry dataset may be valuable, but it is not directly \nisq-ready. Conversely, a compact impedance dataset with well-defined labels may be suitable for quantum-kernel or variational-head experiments. IonSense-QKG introduces a dataset-level quantum-readiness layer to make such distinctions explicit.

\section{Source Corpus and Artifact}
\label{sec:corpus}

EV-Battery-IonSense is a curated open-source intelligence hub for lithium-ion and EV battery monitoring, degradation research, and AI-powered health estimation~\cite{ionsense}. It indexes public battery datasets, source links, related publications, and structured categories. The repository acts as a navigational index and does not host or redistribute original datasets. This is important for licensing and attribution: each dataset remains the property of its original publisher, and users are directed to cite and follow the terms of the original sources.

For IonSense-QKG, the repository is used as the \emph{seed corpus}. The
artifact now contains (i) a quantum-readiness schema, (ii) seed, enriched,
and ranked metadata tables, (iii) a deterministic QRS computation script,
(iv) robustness outputs for the scoring variants, (v) a SQL query workload,
and (vi) link-checking and contribution utilities. The prototype does not
claim complete coverage of all public battery datasets; rather, it provides
a structured path from a curated index to a queryable quantum-readiness layer.

\begin{figure*}[!t]
\centering
\begin{tikzpicture}[
  font=\footnotesize,
  node distance=7mm and 8mm,
  >=Latex,
  stage/.style={
    draw,
    rounded corners=2pt,
    align=center,
    minimum height=10mm,
    text width=2.5cm,
    fill=gray!8
  },
  output/.style={
    draw,
    rounded corners=2pt,
    align=center,
    minimum height=10mm,
    text width=2.9cm,
    fill=gray!16
  },
  labelbox/.style={
    align=center,
    font=\scriptsize
  },
  note/.style={
    align=center,
    text width=3.8cm
  },
  flow/.style={->, thick},
  aux/.style={->, dashed, thick}
]

\node[stage] (repo) {EV-Battery-\\IonSense\\seed index};
\node[stage, right=of repo] (meta) {Quantum-ready\\metadata\\enrichment};
\node[stage, right=of meta] (kg) {IonSense-QKG\\metadata layer};
\node[stage, right=of kg] (qrs) {Weighted QRS\\ranking};
\node[output, right=of qrs] (query) {Discovery queries\\and candidate datasets};

\draw[flow] (repo) -- (meta);
\draw[flow] (meta) -- (kg);
\draw[flow] (kg) -- (qrs);
\draw[flow] (qrs) -- (query);

\node[labelbox, above=1.5mm of repo] {Corpus};
\node[labelbox, above=1.5mm of meta] {Enrichment};
\node[labelbox, above=1.5mm of kg] {Representation};
\node[labelbox, above=1.5mm of qrs] {Scoring};
\node[labelbox, above=1.5mm of query] {Retrieval};

\node[note, below=6mm of meta] (meta_note)
{task, modality, sequence type,\\
label type, access status,\\
preprocessing, encodings};

\draw[aux] (meta_note.north) -- (meta.south);

\node[note, below=6mm of qrs] (qrs_note)
{feature compactness, sequence suitability,\\
modality compatibility, label availability,\\
preprocessing feasibility, reproducibility};

\draw[aux] (qrs_note.north) -- (qrs.south);

\end{tikzpicture}
\caption{IonSense-QKG workflow. A curated battery dataset index is enriched with quantum-relevant metadata, represented as a KG-compatible metadata layer, ranked using a weighted Quantum Readiness Score (QRS), and queried to identify candidate datasets for future hybrid quantum--classical benchmarking.}
\label{fig:pipeline}
\end{figure*}
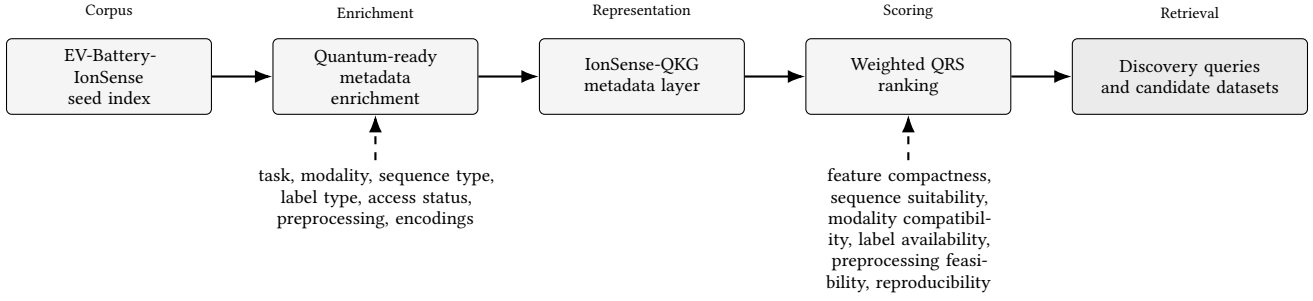

Representative examples include multi-year field measurements with voltage, current, power, and temperature time-series; BMS cloud datasets with rare critical failure classes; early cycle-life prediction datasets; EIS and impedance datasets; voltage relaxation datasets; battery failure and abuse-test resources; CT/X-ray/ultrasound imaging datasets; and NASA battery prognostics datasets. These examples span the range from highly \nisq-feasible compact tabular/spectral datasets to large-scale field and imaging datasets requiring substantial preprocessing.

\section{IonSense-QKG Design}
\label{sec:qkg}

\subsection{Entity model}
IonSense-QKG represents battery datasets and quantum-readiness metadata as linked entities. The graph uses relationships such as \texttt{hasTask}, \texttt{hasModality}, \texttt{hasChemistry}, \texttt{hasLabelType}, \texttt{hasSequenceType}, \texttt{requiresPreprocessing}, \texttt{supportsEncoding}, \texttt{hasAccessStatus}, \texttt{citedBy}, and \texttt{hasNISQFeasibility}. The core entities are listed in Table~\ref{tab:entities}. This representation enables both human-readable browsing and machine-queryable dataset selection. The current artifact realises this model as CSV metadata and SQL-style
queries; a native RDF/SPARQL or graph-database backend is left for future
work.

\begin{table}[!htbp]
  \caption{Core entities in IonSense-QKG.}
  \label{tab:entities}
  \begin{tabularx}{\columnwidth}{@{}p{0.35\columnwidth}Y@{}}
    \toprule
    Entity & Description \\
    \midrule
    \texttt{Dataset} & A public battery dataset or repository \\
    \texttt{Paper} & Publication associated with a dataset \\
    \texttt{Task} & SoH, SoC, RUL, anomaly, failure, capacity, degradation \\
    \texttt{Modality} & BMS time-series, EIS, OCV, voltage relaxation, imaging, strain, thermal, magnetic \\
    \texttt{Chemistry} & LFP, NMC, NCA, LCO, mixed, unknown \\
    \texttt{Label} & Continuous, class, anomaly/failure event, weak, missing \\
    \texttt{SequenceType} & Tabular, time-series, spectral, curve, image, video, graph, simulation \\
    \texttt{AccessStatus} & Public, restricted, code-only, unavailable \\
    \texttt{EncodingCandidate} & Angle, amplitude, basis, quantum kernel, reservoir, variational head \\
    \texttt{NISQFeasibility} & High, medium, low \\
    \bottomrule
  \end{tabularx}
\end{table}

\subsection{Metadata record and seed annotations}
Each dataset entry is enriched with a structured metadata record. A minimal example is shown below.
\begin{verbatim}
dataset_id: wmg_dib_eis
name: WMG-DIB EIS Dataset
tasks: [SoH estimation, capacity estimation]
modalities: [EIS, temperature, SOC-conditioned]
sequence_type: spectral/tabular
label_type: continuous SoH
preprocessing_required: [feature extraction, normalisation]
quantum_encoding_candidates:
  [angle encoding, quantum kernel, variational head]
estimated_qubit_range: 4-8
nisq_feasibility: high
access_status: public
\end{verbatim}

Representative seed annotations are released in the artifact and include
dataset-level task signals, modalities, candidate quantum encodings,
estimated qubit ranges, preprocessing notes, and feasibility labels. These
annotations are used to support query-based dataset selection and QRS
ranking, but they are not empirical quantum benchmark results.

\section{Quantum Readiness Score}
\label{sec:qrs}

\subsection{Design principle}
The Quantum Readiness Score (QRS) is a dataset-selection heuristic
rather than a proof of quantum advantage. It estimates how suitable
a dataset is for near-term hybrid quantum ML experiments. A high
score indicates compact or compressible input features, clear labels,
manageable sequence length, public access, high preprocessing feasibility,
relevance to battery-health tasks, plausible encoding using 4--12 qubits,
and potential for fair classical-vs.-quantum benchmarking.

We define:
\begin{equation}
\mathrm{QRS}(D)=\sum_{i\in\{f,s,m,l,p,a\}} w_i C_i(D),
\label{eq:qrs}
\end{equation}
where the criteria $C_i(D)$ correspond to feature compactness and
estimated qubit feasibility, sequence suitability, modality compatibility,
label availability, preprocessing feasibility, and access/reproducibility
readiness. For this workshop version, the weights are expert-defined
and transparent. Later versions can validate or learn these weights from
benchmark outcomes or user feedback.

Table~\ref{tab:qrsrubric} defines the discrete component scores used by
the QRS formulation.
\begin{table*}[!t]
  \caption{QRS component rubric. Scores are feasibility labels for NISQ workflow readiness, not empirical quantum-performance metrics.}
  \label{tab:qrsrubric}
  \centering
  \footnotesize
  \setlength{\tabcolsep}{2.5pt}
  \renewcommand{\arraystretch}{0.90}
  \begin{tabularx}{\textwidth}{@{}p{0.06\textwidth}p{0.17\textwidth}p{0.25\textwidth}p{0.25\textwidth}p{0.22\textwidth}@{}}
    \toprule
    Sym. & Criterion & High / 1.0 & Medium / 0.5 & Low / 0.0 \\
    \midrule
    $F$ & Feature compactness &
    $\leq$16 features or compact curves/spectra &
    16--128 features after preprocessing &
    Raw images, videos, or large streams \\

    $S$ & Sequence suitability &
    Short curves, EIS, relaxation, or cycle summaries &
    Windowable telemetry or lab trajectories &
    Irregular logs or unclear sampling \\

    $M$ & Modality compatibility &
    EIS, impedance, voltage/capacity curves, compact BMS windows &
    Thermal, strain, ultrasound, magnetic, or mechanical signals after features &
    Raw CT, X-ray, video, or voxel data without embeddings \\

    $L$ & Label availability &
    Clear SoH, SoC, RUL, failure, anomaly, or second-life labels &
    Weak, derived, partial, or noisy labels &
    Missing or unclear labels \\

    $P$ & Preprocessing feasibility &
    Cleaning and normalisation sufficient &
    Windowing, aggregation, PCA, or feature extraction required &
    Heavy segmentation, reconstruction, or domain-specific preprocessing \\

    $A$ & Access readiness &
    Public, documented, downloadable, and citable &
    Partially documented or source-data only &
    Restricted, unavailable, or unclear licensing \\
    \bottomrule
  \end{tabularx}
\end{table*}

\subsection{Initial QRS computation}

For the workshop artifact, we instantiate the scoring rubric on an
initial enriched seed set of 15 representative battery dataset resources
selected from EV-Battery-IonSense. The goal of this computation is not
to empirically validate quantum advantage, but to demonstrate that the
proposed metadata schema and QRS rubric can produce a reproducible
first-pass ranking for dataset selection.

Each dataset is assigned six component scores: feature compactness,
sequence suitability, modality compatibility, label availability,
preprocessing readiness, and access/reproducibility. For this initial implementation, each component
is discretised as \emph{high}~=~1.0, \emph{medium}~=~0.5, and
\emph{low}~=~0.0, then combined using the weighted QRS formula:
\begin{equation}
\begin{split}
\mathrm{QRS}(D)=
&\,0.25F(D)+0.20S(D)+0.20M(D)\\
&+0.15L(D)+0.15P(D)+0.05A(D).
\end{split}
\label{eq:qrs_weighted}
\end{equation}
The weights intentionally prioritise quantum-workflow constraints:
feature compactness, sequence suitability, and modality compatibility
jointly account for 65\% of the score because they directly affect qubit
count, circuit depth, and encoding feasibility.

The resulting ranked table is released in the artifact as
\texttt{metadata/datasets\_qkg\_ranked.csv}, while the enriched
component-level metadata is released as
\texttt{metadata/datasets\_qkg\_enriched.csv}. This ranking should be
interpreted as a transparent candidate-selection heuristic for future
hybrid quantum--classical experiments, not as evidence that the
highest-ranked datasets will necessarily yield quantum advantage.

In addition to the primary weighted additive score, the released artifact
computes two robustness variants: a gated score that emphasises hard NISQ
constraints and a continuous metadata-derived score that relaxes the
high/medium/low discretisation.

\paragraph{Robustness check.}
To assess whether the QRS ranking is an artifact of the chosen weights,
we performed an internal robustness check. Under 2000 Gaussian weight
perturbations ($\sigma=0.05$, renormalised), the ranking retained a mean
Spearman correlation of $\rho\approx0.999$ with the base ranking, with all
five top-ranked datasets remaining in the top five in every run. In
cross-method comparison, the gated score produced identical rank agreement
($\rho=1.00$), while the continuous metadata-derived score showed high
agreement ($\rho\approx0.90$). These checks indicate internal consistency
of the annotation scheme, not empirical validation of quantum advantage.

\subsection{Encoding candidates by modality}
Table~\ref{tab:modality_encoding} maps battery modalities to plausible quantum encodings. The mapping is intentionally conservative: raw images and billion-row telemetry are not treated as directly \nisq-ready, but may become feasible after classical compression, windowing, or representation learning.

\begin{table}[!htbp]
  \caption{Battery modalities and candidate quantum encodings.}
  \label{tab:modality_encoding}
  \begin{tabularx}{\columnwidth}{@{}p{0.34\columnwidth}Y@{}}
    \toprule
    Modality & Candidate quantum encoding \\
    \midrule
    EIS / impedance & angle encoding, amplitude encoding, quantum kernel \\
    Voltage relaxation curves & angle encoding, quantum reservoir, PQC head \\
    BMS time-series windows & recurrent/reservoir encoding, data re-uploading \\
    Cycle-summary features & angle encoding, quantum kernel, PQC head \\
    Capacity / differential-capacity curves & curve-shape encoding, quantum kernel \\
    Strain / ultrasound signals & feature extraction plus quantum classifier \\
    CT / X-ray / imaging & classical embedding plus quantum head/kernel \\
    \bottomrule
  \end{tabularx}
\end{table}

\section{Query Workload}
\label{sec:queries}

IonSense-QKG is designed around dataset-discovery queries. The purpose is not to run quantum models on every dataset, but to show that enriched metadata can guide researchers toward feasible and relevant benchmark choices.

\subsection{Q1: NISQ-feasible hybrid QML datasets}
\begin{verbatim}
SELECT dataset_name, modality, task_type,
       estimated_qubits_max, qrs
FROM IonSenseQKG
WHERE access_status = 'public'
  AND estimated_qubits_max <= 8
  AND label_type IS NOT NULL
ORDER BY qrs DESC;
\end{verbatim}
Expected high-ranking datasets include EIS, impedance,
voltage-relaxation, early-cycle prediction, and compact
cycle-summary datasets.

\subsection{Q2: Quantum time-series encoding}
\begin{verbatim}
SELECT dataset_name, task_type, modality, preprocessing_need
FROM IonSenseQKG
WHERE sequence_type LIKE '%time-series%'
   OR preprocessing_need LIKE '%windowing%'
   OR preprocessing_need LIKE '%aggregation%'
   OR preprocessing_need LIKE '%feature extraction%';
\end{verbatim}
This query returns datasets that either contain time-series structure or
require windowing, aggregation, or feature extraction before quantum
encoding. In the current seed artifact, results include compact diagnostic
datasets such as WMG-DIB EIS and Stanford-MIT early-cycle data, lab and
pack time-series such as NASA, PHEV thermal-fault, and WLTP/constant-discharge
resources, and field or sensor-derived datasets such as BMS cloud, OSF
magnetometry, Tsinghua EV charging, BEV energy dynamics, and home-storage
measurements.

\subsection{Q3: Limited-label anomaly and failure detection}
\begin{verbatim}
SELECT dataset_name, modality, label_type, qrs
FROM IonSenseQKG
WHERE task_type LIKE '%fault%'
   OR task_type LIKE '%anomaly%'
   OR task_type LIKE '%failure%'
ORDER BY qrs DESC;
\end{verbatim}
In the current seed artifact, this query returns NREL abuse/failure data,
PHEV thermal-fault data, and BMS cloud fault-diagnosis data. These examples
illustrate how IonSense-QKG can identify limited-label failure and anomaly
resources for future quantum-kernel or compact hybrid-classifier benchmarks.

\section{Initial Findings}
\label{sec:findings}

We applied the \qrs rubric to an initial enriched seed set of 15 representative battery dataset resources. Table~\ref{tab:qrs_representative_robustness} shows the resulting metadata-driven ranking. The ranking is intended as a dataset-selection aid: it prioritises datasets that are compact, label-rich, accessible, and compatible with low-dimensional quantum encodings. It does not imply empirical quantum advantage.

\begin{table*}[!t]
  \caption{Representative datasets under weighted QRS and alternative scoring variants. QRS$_g$ denotes the gated score and QRS$_c$ denotes the continuous metadata-derived score. These variants are internal robustness checks, not empirical quantum-performance metrics.}
  \label{tab:qrs_representative_robustness}
  \centering
  \small
  \setlength{\tabcolsep}{2pt}
  \renewcommand{\arraystretch}{0.92}
  \begin{tabularx}{\textwidth}{@{}r p{0.30\textwidth} p{0.11\textwidth} p{0.07\textwidth} C{0.045\textwidth} C{0.055\textwidth} C{0.075\textwidth} p{0.19\textwidth}@{}}
    \toprule
    Rank & Dataset & Signal & Band & QRS & QRS$_g$ & QRS$_c$ & Interpretation \\
    \midrule
    1 & Impedance-Based Forecasting Dataset & Impedance/EIS & High & 0.925 & 0.900 & 0.763 & Compact diagnostic candidate \\
    2 & PulseBat Retired-Cell Dataset & Pulse response & High & 0.925 & 0.900 & 0.753 & Retired-cell SoH/second-life \\
    3 & Voltage Relaxation Capacity Estimation Dataset & Relaxation curve & High & 0.925 & 0.900 & 0.763 & Curve-based capacity task \\
    4 & WMG-DIB EIS Dataset & EIS spectra & High & 0.925 & 0.900 & 0.763 & Feature-map candidate \\
    5 & Stanford-MIT Early Cycle Life Dataset & Cycle summaries & High & 0.900 & 0.850 & 0.632 & Early-cycle RUL prediction \\
    \midrule
    6 & NASA Li-ion Battery Aging Dataset & V/I/T series & Medium & 0.600 & 0.450 & 0.547 & Windowing required \\
    11 & OSF Battery Magnetometry Archive & Magnetic scans & Medium & 0.525 & 0.400 & 0.555 & Feature extraction needed \\
    \midrule
    14 & Home-Storage Field Measurements & Field telemetry & Lower & 0.500 & 0.375 & 0.462 & Aggregation required \\
    15 & Battery Imaging Library & CT/X-ray & Low & 0.125 & 0.000 & 0.181 & Classical embedding required \\
    \bottomrule
  \end{tabularx}
\end{table*}

Using EV-Battery-IonSense as a seed corpus and the representative QRS results in Table~\ref{tab:qrs_representative_robustness}, IonSense-QKG supports five preliminary observations.

\textbf{Finding 1: NISQ-feasible battery datasets are usually compact diagnostic datasets, not the largest field datasets.} Large-scale field datasets are valuable for real-world validation, but their raw scale makes them less directly suitable for \nisq experiments. In contrast, EIS, impedance, voltage relaxation, early-cycle summary, and controlled ageing datasets often provide compact feature structures suitable for 4--12 qubit hybrid models. Quantum battery benchmarking should therefore begin with diagnostic and cycle-summary datasets before scaling to raw fleet telemetry.

\textbf{Finding 2: Time-series datasets require a representation layer before quantum encoding.} Many EV and BMS datasets provide high-resolution voltage, current, temperature, power, and charging behaviour. These are promising for quantum time-series methods, but direct encoding of long sequences is infeasible on near-term devices. Practical workflows require windowing, aggregation, recurrence summaries, reservoir states, or classical embeddings before quantum processing.

\textbf{Finding 3: Failure and anomaly datasets are promising for limited-label quantum learning.} Battery failure datasets are often small, imbalanced, or rare-event driven. This makes them interesting for quantum kernel methods, compact classifiers, or hybrid models under limited labels.

\textbf{Finding 4: EIS and voltage-relaxation modalities are strong candidates for quantum feature maps.} EIS, impedance, and voltage-relaxation data are naturally structured as curves or spectra. They are often more compact than raw field telemetry or imaging data, while retaining rich electrochemical information. This makes them strong candidates for angle encoding, amplitude encoding, quantum kernels, and variational quantum heads.

\textbf{Finding 5: Imaging data is scientifically rich but not directly NISQ-ready.} Battery CT, X-ray, synchrotron, and ultrasonic datasets provide valuable structural and diagnostic information. However, raw image or voxel data is too high-dimensional for direct \nisq encoding. A more realistic workflow is classical feature extraction or representation learning followed by a quantum kernel/classifier/head.

\section{Prototype Implementation}
\label{sec:prototype}

IonSense-QKG is implemented as a lightweight, repository-level artifact rather than as a redistribution of battery data. The prototype has three layers. The \textbf{metadata extraction layer} uses EV-Battery-IonSense as the seed corpus and extracts dataset names, categories, scale descriptions, access links, and related papers. The \textbf{manual/assisted enrichment layer} adds quantum-relevant fields such as modality, task, sequence type, label type, estimated qubit range, preprocessing requirement, candidate quantum encoding, and \nisq feasibility. The \textbf{query and ranking layer} stores enriched metadata as CSV and computes a deterministic \qrs ranking using the rubric in Section~\ref{sec:qrs}.

The current public artifact implements this first lightweight version. It includes the quantum-readiness schema, an initial seed metadata table, an enriched metadata table, a ranked QRS output, a link-checking script, a deterministic QRS computation script, robustness outputs, example discovery queries, and contribution guidelines. The artifact is organised in the public repository under \texttt{metadata/}, \texttt{scripts/}, and \texttt{examples/}, with the ranked metadata, QRS computation script, robustness report, and query workload released alongside the schema.

The prototype does not redistribute original battery datasets. It provides curated links, quantum-readiness metadata, a deterministic QRS computation script, a ranked metadata table, and example dataset-discovery queries. This keeps the artifact lightweight while still making the proposed metadata schema and QRS ranking reproducible.

\section{Discussion}
\label{sec:discussion}

\subsection{Relation to quantum data/knowledge management}
IonSense-QKG contributes to quantum data and knowledge management by treating quantum ML dataset selection as a queryable metadata problem. Rather than asking whether a particular quantum model outperforms a classical model, the framework asks which datasets are appropriate for different quantum workflow designs. This aligns with data-centric quantum research, where progress depends not only on algorithms but also on benchmark selection, metadata quality, provenance, and reproducibility.

\subsection{Justification and interpretation of QRS}

The proposed Quantum Readiness Score is not intended to be a ground-truth measure of quantum advantage. At present, no public benchmark provides empirical labels indicating which battery datasets are truly favourable for hybrid quantum--classical learning. Therefore, QRS should be interpreted as a transparent dataset-selection heuristic rather than as a validated performance predictor.

The justification for QRS is constraint-based. Near-term hybrid quantum workflows are limited by input dimensionality, circuit depth, shot budgets, noise, and the availability of meaningful labels. The six QRS components therefore correspond to practical constraints that directly affect whether a dataset can be used in a NISQ-era workflow: feature compactness controls the number of encoded variables; sequence suitability captures whether time-series data can be windowed or compressed; modality compatibility reflects whether the signal type naturally supports quantum feature maps; label availability determines whether supervised or evaluation tasks are possible; preprocessing feasibility captures the cost of producing quantum-ready inputs; and access/reproducibility determines whether results can be independently reproduced.

This makes QRS useful as a first-pass filtering and prioritisation tool. A high QRS does not imply that a quantum model will outperform classical baselines. It only indicates that the dataset is likely to be feasible for near-term hybrid quantum experimentation. Conversely, a low QRS does not mean that a dataset is scientifically unimportant; it may simply require stronger classical preprocessing, feature extraction, or embedding before quantum processing. Empirical validation of QRS through downstream quantum benchmarks is left as future work.
\subsection{Avoiding overlap with dataset-descriptor work}
This work is not a new battery dataset and does not duplicate any dataset descriptor submission. It does not generate new battery signals, host original data, or claim ownership over source datasets. Its contribution is a KG-compatible quantum-readiness metadata layer over an existing public index. This separation is important for avoiding overlap with dataset descriptor papers and for keeping the contribution aligned with data-centric battery and quantum-workflow research.

\subsection{Limitations and future work}
The current framework has several limitations. First, QRS is an expert-defined, metadata-driven heuristic and does not have ground-truth labels for validation. It should therefore be interpreted as a candidate-selection score rather than as a predictor of quantum advantage or downstream accuracy. Second, some metadata fields require manual annotation because dataset descriptions are inconsistent. Third, access status and licensing may change over time. Fourth, estimated qubit requirements depend on preprocessing choices and model design. Finally, future work should validate the QRS ranking by running selected high-, medium-, and low-readiness datasets through common hybrid quantum pipelines and measuring whether QRS correlates with feasibility, runtime, reproducibility, and downstream performance.

\section{Conclusion}
\label{sec:conclusion}

This paper introduced IonSense-QKG, a quantum-readiness metadata framework for lithium-ion battery dataset discovery. Starting from EV-Battery-IonSense, the proposed framework enriches public battery dataset metadata with quantum-relevant annotations, including task type, modality, sequence structure, label availability, preprocessing requirements, candidate quantum encodings, estimated qubit feasibility, and a Quantum Readiness Score. Through a set of dataset-discovery queries, IonSense-QKG helps identify which public battery datasets are feasible for \nisq-era hybrid QML, which datasets support quantum time-series encoding, which resources are suitable for limited-label anomaly/failure detection, and which modalities are most promising for quantum feature maps. The work positions dataset selection as a knowledge-management problem and provides a practical step toward reproducible, data-centric quantum battery analytics.

\begin{acks}
The authors acknowledge the maintainers and original publishers of the public battery datasets indexed by EV-Battery-IonSense. This work does not redistribute original datasets; users should cite and follow the terms of each source dataset. Any errors in metadata interpretation or quantum-readiness annotation remain the responsibility of the authors.
\end{acks}

\end{document}